\documentclass[lettersize,journal]{IEEEtran}
\usepackage{amsmath,amsfonts}
\usepackage{algorithmic}
\usepackage{algorithm}
\usepackage{array}
\usepackage[caption=false,font=normalsize,labelfont=sf,textfont=sf]{subfig}
\usepackage{textcomp}
\usepackage{stfloats}
\usepackage{url}
\usepackage{verbatim}
\usepackage{graphicx}
\usepackage{cite}
\usepackage{booktabs}

\begin{document}

\title{Intelligent Cross-Organizational Process Mining: A Survey and New Perspectives}

\author{Yiyuan Yang, Zheshun Wu, Yong Chu, Zhenghua Chen, Zenglin Xu$^\dag$, Qingsong Wen$^\dag$
\thanks{Yiyuan Yang is with the Department of Computer Science, University of Oxford, UK (yiyuan.yang@cs.ox.ac.uk).}
\thanks{Zheshun Wu and Yong Chu are with the Peng Cheng Laboratory, China (wuzhsh23@gmail.com and chuyong.edu@gmail.com).}
\thanks{Zhenghua Chen is with Institute for Infocomm Research, A*STAR, Singapore (chen0832@e.ntu.edu.sg).}
\thanks{Zenglin Xu is with Fudan University, China (zenglin@gmail.com).}
\thanks{Qingsong Wen is with Squirrel Ai Learning, USA (qingsongedu@gmail.com).}
\thanks {$^\dag$Corresponding authors: Zenglin Xu and Qingsong Wen. \\Version date:  \today.}
}

\markboth{Journal of \LaTeX\ Class Files,~Vol.~14, No.~8, August~2021}%
{Shell \MakeLowercase{\textit{et al.}}: A Sample Article Using IEEEtran.cls for IEEE Journals}

\maketitle

\begin{abstract}
Process mining, as a high-level field in data mining, plays a crucial role in enhancing operational efficiency and decision-making across organizations. In this survey paper, we delve into the growing significance and ongoing trends in the field of process mining, advocating a specific viewpoint on its contents, application, and development in modern businesses and process management, particularly in cross-organizational settings. We first summarize the framework of process mining, common industrial applications, and the latest advances combined with artificial intelligence, such as workflow optimization, compliance checking, and performance analysis. Then, we propose a holistic framework for intelligent process analysis and outline initial methodologies in cross-organizational settings, highlighting both challenges and opportunities. This particular perspective aims to revolutionize process mining by leveraging artificial intelligence to offer sophisticated solutions for complex, multi-organizational data analysis. By integrating advanced machine learning techniques, we can enhance predictive capabilities, streamline processes, and facilitate real-time decision-making. Furthermore, we pinpoint avenues for future investigations within the research community, encouraging the exploration of innovative algorithms, data integration strategies, and privacy-preserving methods to fully harness the potential of process mining in diverse, interconnected business environments.
\end{abstract}


\section{Introduction}
\IEEEPARstart{A}{rtificial} intelligence (AI) represents one of the most recent frontiers in science and engineering~\cite{russell2010artificial}. AI technology is widely utilized across industry, government, and scientific sectors, with prominent applications including recommendation systems~\cite{tavakoli2022ai}, natural language understanding~\cite{wilks1976natural}, self-driving technology~\cite{hossain2019caias}, as well as superman analysis in strategy games~\cite{granter2017alphago}, and business processes~\cite{DEWEERDT201357}. The global AI market size was valued at USD 454 billion in 2022 and is expected to hit around USD 2,575 billion by 2032, progressing with a compound annual growth rate (CAGR) of 19\% from 2023 to 2032\footnote{https://www.precedenceresearch.com/}.

From an industrial perspective, process mining is increasingly recognized as a crucial tool for enhancing operational efficiency, compliance, and overall business performance~\cite{van2012process}.
Process mining is a technique used to analyze business processes based on event logs extracted from a company’s information systems. It provides an objective, data-driven view of how processes are actually performed, distinguishing it from traditional process modeling, which is often based on subjective views and assumptions. In consequence, process mining techniques for various industries have been expanding quickly and grabbing the interest of worldwide investors and enterprises. 

\textbf{Why This Paper?} Traditional process mining methodologies, which typically depend on data from a single organizational source to uncover or improve business processes, often fall short in addressing the complexities of more intricate applications. This limitation can be overcome by leveraging the vast amounts of data residing in distributed database systems across various organizations, thereby significantly improving the effectiveness of process mining. This pressing requirement underscores the emergence of a new research domain: \textbf{Cross-Organizational Process Mining}. However, this advancement faces notable obstacles, including privacy concerns, customized process, fast responses, and issues related to data quality, which hinder the progress of cross-organizational process mining initiatives. In response to these challenges, some studies have sought to incorporate techniques from distributed databases into process mining. This approach, however, involves the direct amalgamation of private data, creating a conflict with the tenets of privacy-preserving computing. On a different note, works by researchers such as \cite{khan2021cross}  have introduced concepts like Federated Learning (FL) to process mining, aiming to safeguard user privacy. Yet, these initiatives do not address the challenges posed by the heterogeneous nature of data distributions inherent in FL, nor do they tackle critical aspects such as ensuring data quality and enabling customizable computations. This gap in research underscores a significant barrier to the practical implementation of these advanced process mining techniques in the industrial sector.

Given these circumstances, we introduce the concept of \textbf{Intelligent Cross-Organizational Process Mining} from an industrial perspective. We provide a comprehensive framework for conducting intelligent process mining across various organizations while preserving privacy. The main contributions of this paper are as follows: (1) We present a general overview of typical process mining, along with related industrial applications, methods, metrics, resources, and open-source tools. (2) We discuss a range of cutting-edge techniques and developments designed for process mining, incorporating new types of process mining methods. (3) We introduce a new research direction called intelligent cross-organizational process mining, and propose corresponding initial solutions with a comprehensive analysis. (4) We discuss the challenges and opportunities for intelligent cross-organizational process mining, primarily focusing on practical industry implications.

The remainder of this paper is structured as follows. Section II introduces the fundamental concepts and framework of typical process mining. In Sections III and IV, we delve into the application of advanced data analytics and machine learning technologies within this domain, incorporating methods, metrics, resources, and open-source tools. Section V discusses the proposed intelligent cross-organizational process mining from different perspectives. Section VI highlights the existing challenges in deploying intelligent cross-organizational process mining solutions and suggests potential research directions to address these challenges. Lastly, Section VII provides a conclusion that summarizes the key points discussed and the significance of the advancements in intelligent cross-organizational process mining.

\begin{figure*}[!ht]
    \centering
    \includegraphics[width=0.95\textwidth]{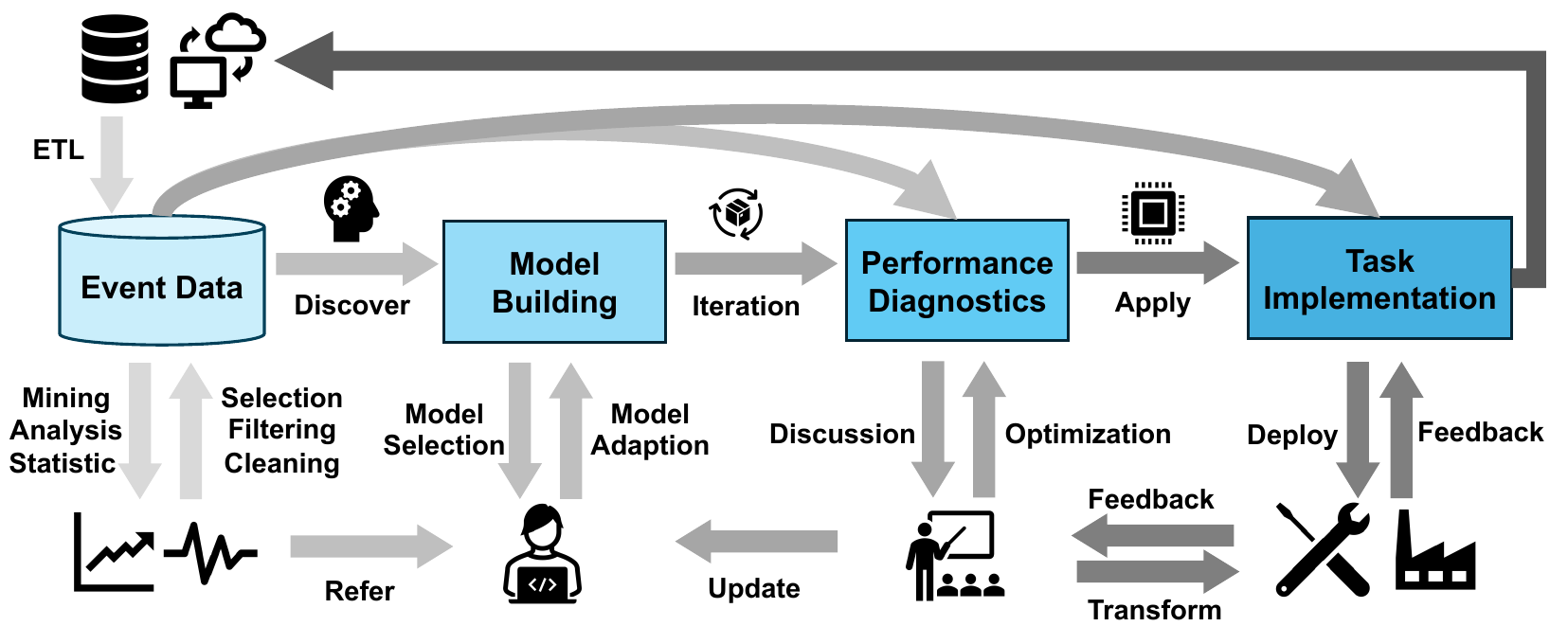}
    \caption{The Workflow of a Typical Process Mining System.}
    \label{fig:workflow}
\end{figure*}

\section{Process Mining}
\subsection{Workflow of Process Mining}
Process mining is a method of extracting structured, interpretable processes from the data stored from business events~\cite{van2004process}. The core of process mining is extracting information from event logs in management systems to discover, monitor, and improve real processes~\cite{van2012process}. Fully mining and reasonably using this information can enhance the operational efficiency of business automation and reduce corporate operational costs. 

Fig.~\ref{fig:workflow} presents the typical workflow of process mining. Specifically, the system first extracts and transforms valid logs from information systems, such as company servers and cloud clients, and loads them into a database as event data. This process is commonly known and abbreviated as ETL (extract, transform, and load). These event data record the running time and associated data for each phase of the process activities, which encompasses the actual flow of business activities. Next, before building a process model, those logs will be mined, analyzed, and conducted preliminary statistics, and based on the results, select data and filter out erroneous samples. When building a process model, heuristic or AI-based structures can be chosen. Spectacular engineers have the task of carefully selecting and tuning models for adaptation. After multiple optimization iterations, the model will be used for performance diagnosis, to test consistency between event logs and the process model, to identify problems in the real process, to analyze causes and to optimize and provide feedback~\cite{rosa2017business}. Finally, execute and deploy the process model and observe its features to assess the system's high performance. Feedback is then provided to the companies, along with relevant log storage.

\subsection{Discovery in Process Mining}
Discovery in process mining refers to the method of automatically discovering business processes from existing event data. In practical applications, the business processes of most organizations have not been completely recorded or understood, especially when these processes are complex, informal, or subject to frequent changes. The discovery process reveals the actual processes by analyzing event logs, which are usually time-series data on business activities recorded by systems. 

Currently, numerous algorithms have been proposed, establishing a bridge between event logs and process models. For example, $\alpha$ miner takes event logs as input and outputs a Petri Net~\cite{van2004workflow}. Based on the event logs, it models the sequential relationships (footprint) between events, including four basic relationships: directly follows $>{L}$, causal $\rightarrow{L}$, unrelated $\#{L}$, and parallel $||{L}$. It is straightforward and allows one to construct process models entirely based on event logs by understanding the relationships and causality between process steps. Additionally, it can handle the discovery of concurrency in processes. However, due to its limited performance in special circumstances, including dealing with noise, infrequent, invisible, and duplicate tasks, complex routing structures (such as short loops), and without considering event frequency or capturing long-distance dependencies, $\alpha$ miner is not practical in real-world applications.

Based on the $\alpha$ miner, several improvement methods have been proposed. The heuristic miner utilizes the frequency information and can handle short loops of lengths 1 and 2. In addition, it can deal with complete dependence, independence, and unobservable tasks, as well as managing long-distance dependencies. However, it requires setting thresholds to determine connections, making it less effective in handling unusual paths~\cite{weijters2003rediscovering}. Fodina Miner exhibits robustness to noise, can identify repetitive activities, and allows user intervention to optimize the discovery process. Therefore, it avoids certain types of deadlocks within the heuristics miner. However, when applied to event logs in real-life scenarios, the models generated by Fodina Miner tend to be large and unreliable~\cite{vanden2017fodina}. The genetic miner extracts features from a global search and addresses noise issues using a genetic algorithm, which finds models with desired fitness through initialization, selection, reproduction, and iteration. The main challenge of it is to define an effective fitness measure, as it guides the global search. Overall, the genetic miner is robust and capable of handling noise and incompleteness, but due to model evaluation metrics, it can result in slow model iteration and may even fail to yield satisfactory models~\cite{van2005genetic}. Building upon the genetic miner, the evolutionary miner is introduced. The main distinction between them is using process trees instead of causal dependency nets. Fitness calculations are also conducted on the process trees, and four quality dimensions are comprehensively considered to better discover process models~\cite{marquez2017run}. The fuzzy miner integrates process mining with clustering, introducing unstructured data into clustering for processing. This approach enhances the relevance of events with high syntactic or semantic similarity, thereby improving performance and robustness. However, the fuzzy miner is based on the ProM framework and cannot be converted into other types of process modeling languages, such as business process model and notation (BPMN) or business process execution language (BPEL). Additionally, the fuzzy miner method does not fully guarantee the rationality of the models~\cite{gunther2007fuzzy}.

\subsection{Conformance in Process Mining}
The purpose of conformance is to assess the quality of the process models, identify problems in real processes, and analyze the causes~\cite{van2016data}. Specifically, conformance helps organizations understand the actual execution of their business processes and identify areas that require improvement. The conformance process is typically divided into the following steps and components~\cite{van2012process,van2004workflow}. (1) Process model: This is the basis for conformance. A process model can be automatically constructed from log data through the discovery part, or a predefined model can be used. (2) Log data: Collection of real process event logs. These logs record every step and activity of the process. (3) Comparative analysis: Use process mining tools to compare the process model with real log data, identifying cases where the process model is violated. (3) Deviation identification: The identified deviations could be unfollowed steps, additional steps, sequence errors, time delays, etc. (4) Performance and efficiency assessment: This step involves evaluating the overall efficiency and performance of the process, such as processing time, waiting time, resource utilization, etc. (5) Reporting and improvement: Generate reports that provide detailed information on process deviations and potential points of improvement, followed by a thorough iterative optimization.

\subsection{Enhancement in Process Mining}
Enhancement involves the use of relevant information to expand a process model, thereby improving or enhancing the performance and efficiency of existing processes~\cite{van2012process}. Process mining models enhance efficiency by optimizing process models by using timestamps in event logs, using AI-based methods and statistical data. In the analysis of the model, process enhancement mainly considers aspects such as average throughput time per case, transition and residence times, process bottlenecks, most critical activities and resources, working and waiting times, etc~\cite{van2016data}. Additionally, enhancement in industrial applications typically involves the following aspects: (1) Optimizing process performance: By analyzing event logs to identify bottlenecks or inefficient steps and proposing improvements. (2) Predicting future trends: Using process mining techniques to predict the future behavior of processes, such as potential delays or bottlenecks, to take proactive measures. (3) Improving decision making: Supporting more informed business decisions by analyzing process data, for example, in resource allocation or priority setting. (4) Compliance and risk management: Monitoring processes for adherence to relevant laws and policies, identifying potential risks and non-compliant behaviors.

\section{Latest Advances and Ideas in Process Mining}
\subsection{Data-centric Process Mining}
For data-centric process mining, data privacy is one of the most significant issues. As commonly understood, privacy refers to the capability of an individual or group to isolate themselves or their information and selectively disclose it~\cite{sun2014data}. Specifically, data privacy encompasses the right to control the collection and utilization of personal information. It represents an individual's or group's capacity to prevent information about themselves from being disclosed to individuals other than those to whom they intentionally provide it. A significant user privacy concern involves the identification of personal information during transmission over the Internet~\cite{porambage2016quest}.  Within organizations, privacy encompasses the application of laws, mechanisms, standards, and processes that govern the management of personally identifiable information~\cite{sun2014data}. In the realm of cloud computing, privacy denotes the capability of cloud services to thwart potential adversaries from deducing a user's behavior based on their visit model when accessing sensitive data~\cite{6187862,mehmood2016protection}. There are some prevalent privacy-preserving methods in big data. De-identification, including K-anonymity and L-diversity, is one of the most important privacy protection techniques~\cite {machanavajjhala2007diversity,li2006t}. The core idea of these methods is to first sanitize the data through processes like generalization (substituting quasi-identifiers with less specific yet semantically consistent values) and suppression (withholding certain values entirely) prior to the release for data mining. Federated learning represents a promising distributed machine learning framework utilized in privacy-sensitive scenarios. Protecting user privacy involves participating clients collaboratively training machine learning models without directly sharing their collected raw data~\cite{zhang2021survey,kairouz2021advances}.

Additionally, data missing presents another critical challenge in data-centric process mining. Missing data, or missing values, occur when no data value is stored for the variable in an observation~\cite{allison2010missing}. There are several mechanisms that can lead to missing data~\cite{qian2024unveiling}. Firstly,  if data is missing and is unrelated to the observed characteristics of the samples, it is termed as missing completely at random (MCAR). Secondly, when the missing data is related to the observed characteristics of the samples, it is termed as missing at random (MAR). Lastly, when the missing data is related to unobserved characteristics of the sample, it is called missing not at random (MNAR)~\cite{guan2011missing}. Broadly, there are some primary approaches to handling missing data: imputation~\cite{woods2021best,du2024tsi}, interpolation~\cite{derrick2017test}, full analysis~\cite{chechik2006max}, model-based techniques~\cite{mirkes2016handling,du2023saits} and so on.

\subsection{Object-centric Process Mining}
Upon extracting an event log from an information system, the resultant log may demonstrate convergence (where one event is related to multiple cases) and divergence (independent, repeated executions of a group of activities within a single case). These occurrences can lead to event replication, potentially resulting in misleading outcomes~\cite{van2019object}.  Real-life processes frequently involve numerous one-to-many and many-to-many relationships, posing challenges for process mining. Object-centric process mining techniques are designed to specifically address these convergence and divergence issues~\cite{9576886,van2019object,li2018extracting}. 

Some related works have tried to address these challenges in object-centric process mining. The authors in ~\cite{van2019object} introduce a specific logging format that allows events to be linked to objects of diverse types. In~\cite{9576886}, the authors present a concept for evaluating the precision and fitness of an object-centric petri net in relation to an object-centric event log.  They further provide a formal definition along with an illustrative example and introduce an algorithm for computing these quality measures. Additionally,~\cite{van2023object}  presents object-centric event data (OCED) and demonstrates how these data can be utilized to uncover, analyze, and enhance the structure of complex real-world processes.

\subsection{Customized Metric and AutoML}
Intelligent process mining involves offering customized computing services tailored to specific users or silos. For instance, distinct users may have varying time prediction tasks, differing in aspects such as data type or prediction horizon. Hence, it is essential for intelligent process mining to automatically adapt the corresponding learning strategies.

Automated machine learning (AutoML), a key machine learning technique, has significantly contributed to the development of customized metrics. With the growth of datasets and computing resources, machine learning techniques have become deeply ingrained in our daily routines. However, achieving optimal learning performance demands substantial knowledge and effort, resulting in extensive human involvement across all aspects of machine learning. In an effort to streamline the application of machine learning techniques and reduce the dependence on experienced human experts, AutoML has emerged as a significant area of interest in both industrial and academic domains~\cite{yao2018taking,zoller2021benchmark,tuggener2019automated}. The core idea of AutoML is to regard the problem of automating the process of combined algorithm selection or hyper-parameter tuning (CASH) as an optimization problem~\cite {elshawi2019automated}. More specifically, given a set of machine learning algorithms, a training set and a validation set, the goal of AutoML is to find a tuned algorithm that achieves the highest generalization performance on the validation set. Several typical AutoML methods include meta-learning, neural architecture search (NAS), hyperparameter optimization, and others. The general workflow of these methods involves inputting data, optimization metrics and constraints into the AutoML system,  then the system returns the corresponding optimal machine learning model to users~\cite{elshawi2019automated}.

\subsection{Efficient Computing Infrastructure}
One of the most important efficient computing infrastructures is distributed computing.  Recently, distributed computing has been utilized for large-scale systems, offering several advantages compared to centralized computing~\cite{cristea2010large}. Firstly, it provides computing services with high reliability and fault tolerance. Distributed computing systems can efficiently and reliably operate even if central nodes fail due to communication bottlenecks. Secondly, distributed computing demonstrates high computation speed as the computational load is shared among various computing nodes. Thirdly, these systems are scalable, meaning that adding more computing nodes enhances the corresponding computing capacity~\cite{ng2020survey}. Given these advantages, distributed computing has found application in numerous real-life scenarios such as distributed machine learning systems~\cite{9999556}, and blockchain systems~\cite{zhao2020blockchain}.

Among these applications, distributed machine learning systems play a key role in implementing cross-organizational process mining. As machine learning techniques continue to advance, the pursuit of higher prediction quality and the applicability of machine learning solutions to more complex scenarios demand a substantial amount of training data~\cite{qiu2016survey,peteiro2013survey}. However, the demand for processing training data has outpaced the increase in the computational power of individual machines. Consequently, there is a need to distribute the machine learning workload across multiple machines, thereby transforming centralized systems into distributed ones~\cite{verbraeken2020survey}. In the context of cross-organizational process mining, a distributed machine learning system can harness the massive event log data stored in distributed database systems to train specific models~\cite{khan2021cross}. These models can subsequently offer various computing services for users, including time prediction in process models~\cite{van2011time} and more.

\section{Industrial Applications and Resources}
\subsection{Application Scenarios}
\begin{figure}[!t]
    \centering
    \includegraphics[width=0.45\textwidth]{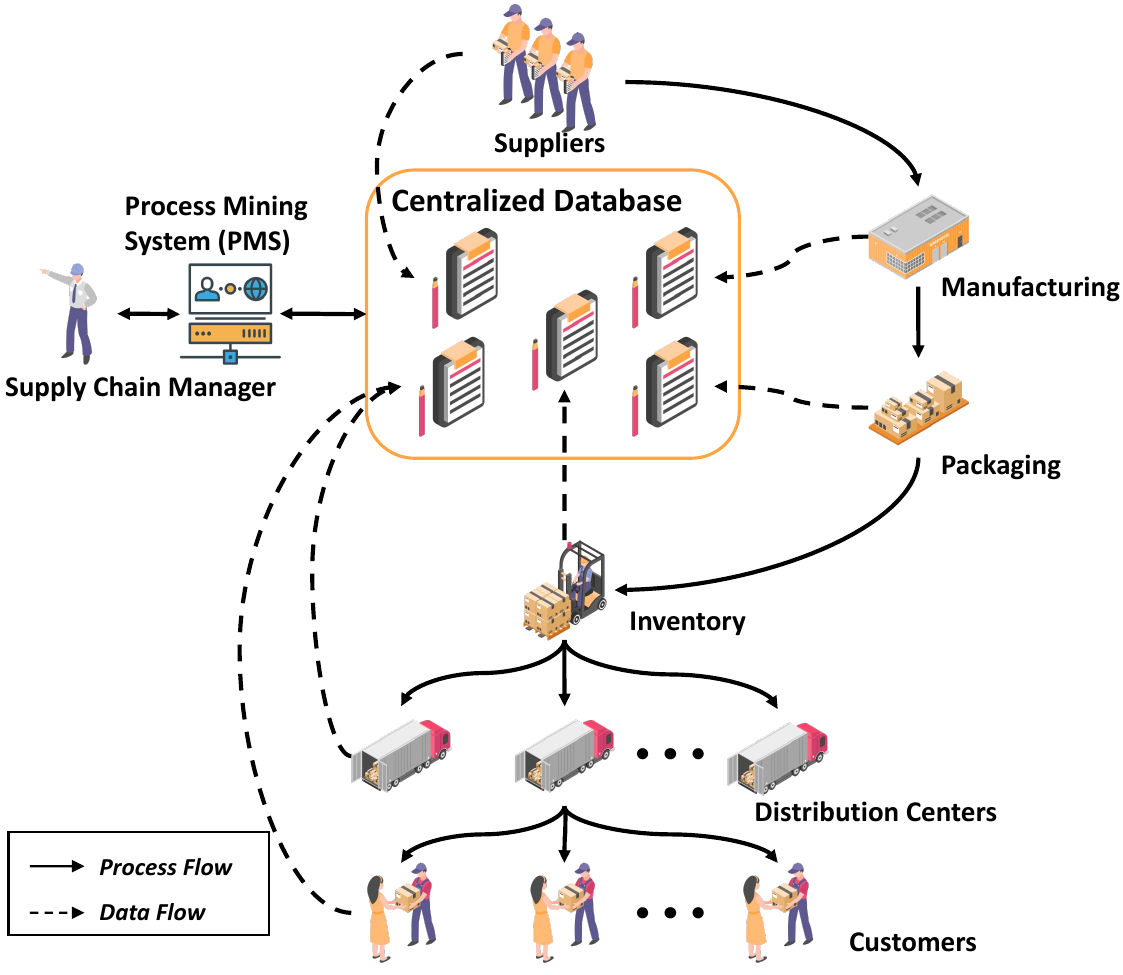}
    \caption{Process Mining for Logistics.}
    \label{fig:logistics}
\end{figure}
\subsubsection{Manufacturing and Logistics} 
In the realm of manufacturing and logistics, business processes are marked by considerable complexity and dynamism. This complexity arises from multiple factors, such as intricate product designs, the involvement of numerous participants, and the occurrence of unexpected events. Therefore, improving the efficiency of business processes is crucial. Fig.~\ref{fig:logistics} illustrates the typical workflow of process mining in logistics processes. Process mining presents a promising approach for gaining insight into existing business processes and has been employed in various studies as part of comprehensive investigations within the manufacturing and logistics sectors~\cite{DBLP:journals/dke/RozinatWAHF09,DBLP:journals/dss/GhattasSP14}. For instance, it has been employed as a foundational element for decision-making in business process management. \cite{DBLP:conf/ldic/IntayoadB18} present a methodology to improve the constraints of process mining. This involves employing a Markov chain as a sequence clustering technique during the data preprocessing stage and utilizing heuristic mining to extract the business process models.

\begin{figure}[!t]
    \centering
    \includegraphics[width=0.485\textwidth]{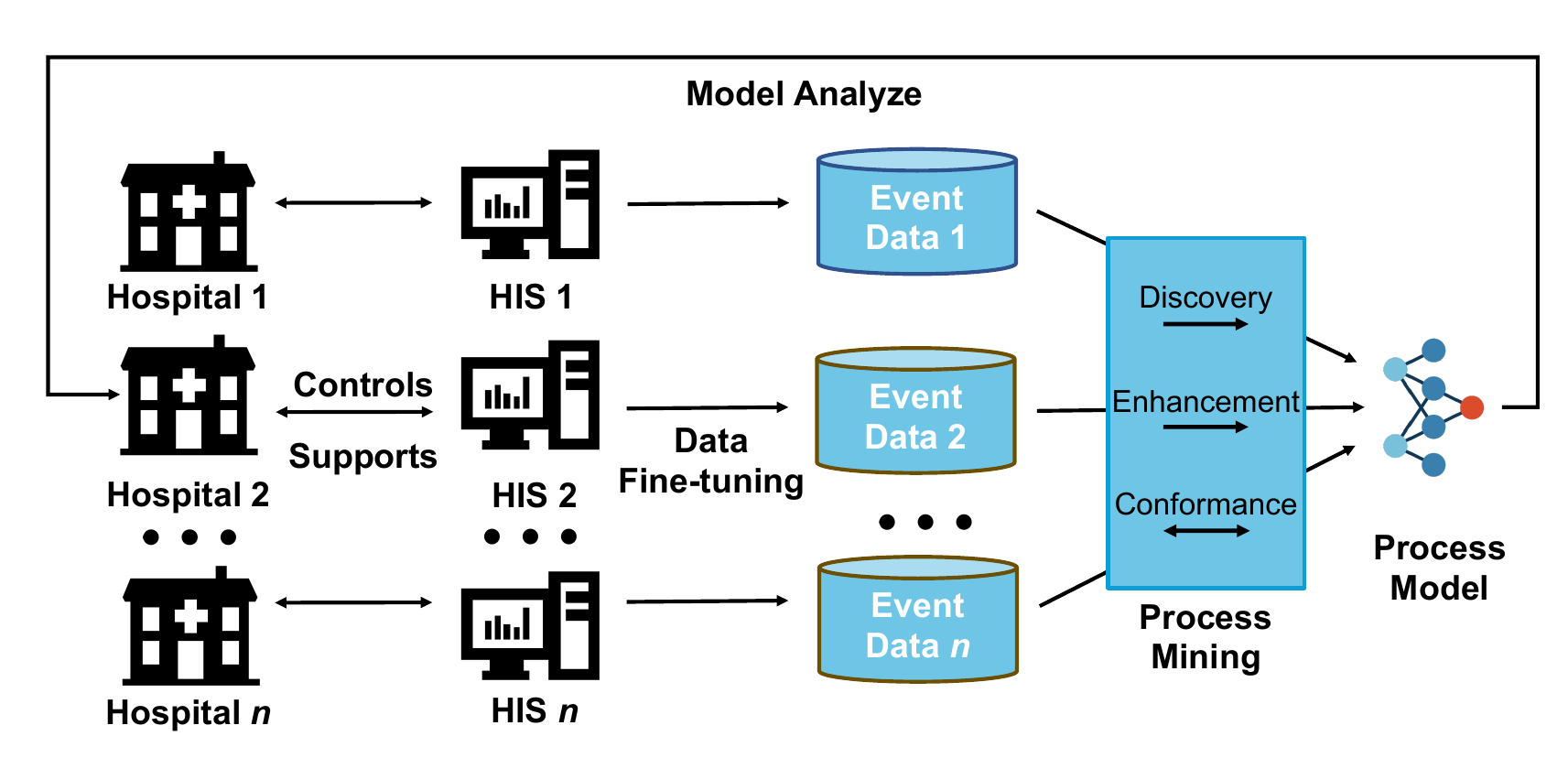} 
    \caption{Process Mining for Healthcare.} 
    \label{fig:healthcare}
\end{figure}
\subsubsection{Healthcare}
Healthcare processes are known for their intricate and ever-changing nature~\cite{6354456}. The application of process mining techniques in healthcare not only aids in a thorough comprehension of these procedures but also leads to advantages related to process efficiency, such as having a positive effect on the management of medical centers. Fig.~\ref{fig:healthcare} presents a broad overview of process mining's application in healthcare. Typically, any action carried out in a hospital by staff to provide care to a patient is recorded in the hospital information system (HIS)~\cite{DBLP:journals/ijeh/GhasemiA16}. These activities are logged as events for assistance and subsequent analysis. Process models are generated to define the sequence where various healthcare workers are expected to carry out their tasks within a specific process. Furthermore, process models are also utilized to aid in the development of HIS.

\begin{figure}[!t]
    \centering
    \includegraphics[width=0.45\textwidth]{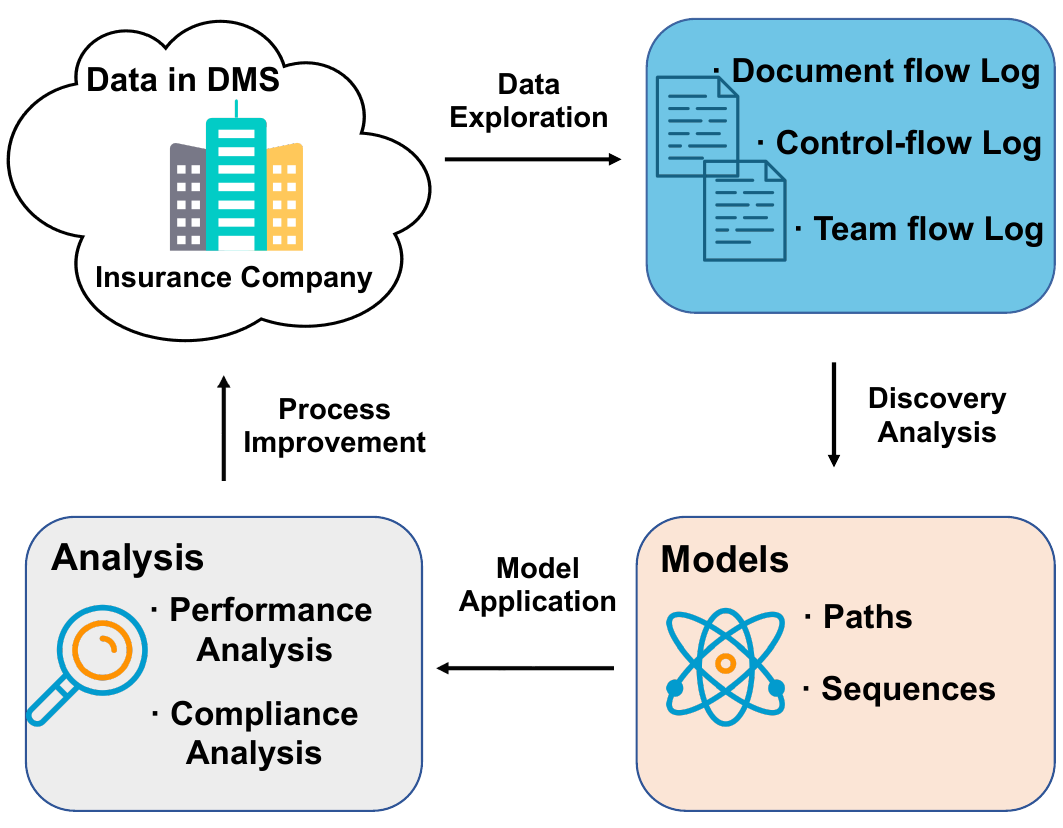}
    \caption{Process Mining for Finance.}
    \label{fig:financial}
\end{figure}
\subsubsection{Finance} 
Process mining has been widely exploited in the financial sector, given that regulatory requirements mean that financial processes yield high volumes of event data. These data are used in financial and compliance auditing ~\cite{financial,2015Hybrid}. Fig.~\ref{fig:financial} presents a methodology framework structuring the process of a real-life analysis and illustrates its usefulness in a multi-faceted case study situated in the financial services industry. The process data was extracted from a document management system implemented by a large Belgian insurance company to support its back office processes~\cite{DEWEERDT201357}. The core information system underlying the insurance company’s back office can be best described as a document management system (DMS). Data in DMS had to be fine-tuned multiple times, going through the scope adjustment loop after each inspection, until a satisfactory data set was obtained. Afterward, the execution data could be decoupled into three different event logs according to three different information perspectives. Models built from these event logs can be used for performance analysis and compliance analysis, thereby identifying repetitions or errors that occur during business processes and improving the entire workflow.

\begin{figure}[!t]
    \centering
    \includegraphics[width=0.45\textwidth]{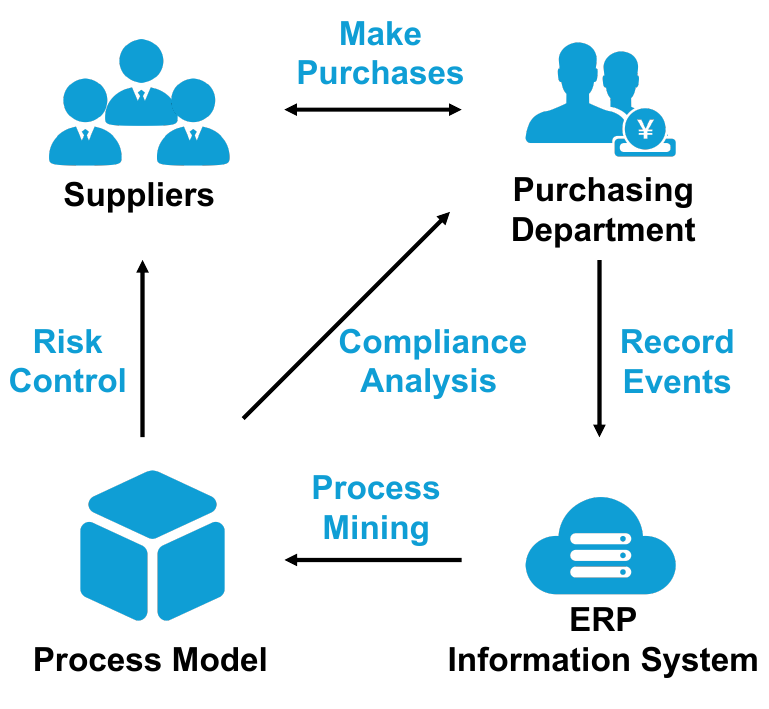}
    \caption{Process Mining for Procurement.} 
    \label{fig:procurement}
\end{figure}
\subsubsection{Procurement}
Effective procurement processes are integral to an organization’s value chain, facilitating the delivery of services and products~\cite{diba2019compliance}. Fig.~\ref{fig:procurement} demonstrates how process mining plays a role in the procurement process. By analyzing these processes, valuable insights can be gained to identify improvement opportunities and mitigate potential risks. Process mining is one of the tools used to analyze the procurement process, offering techniques for scrutinizing processes using recorded data, such as the provided event log. For instance, the authors in~\cite{darzi2023analysis} perform process mining in the procurement process. They begin by extracting raw data from the process-maker software. Subsequently, they analyze the process path by examining the purchase process of the American Productivity and Quality Center (APQC) and extracting the corresponding event log. Process mining enables them to offer valuable insights and recommendations to the organization’s management unit through the application of process mining.

\begin{figure}[!t]
    \centering
    \includegraphics[width=0.485\textwidth]{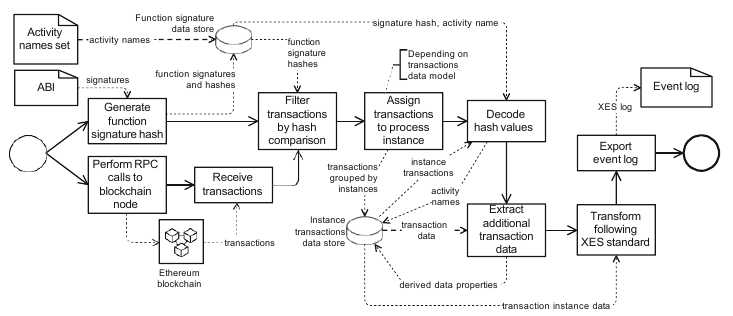}
    \caption{Process mining for Blockchain.} 
    \label{fig:blockchain}
\end{figure}
\subsubsection{Blockchain}
Blockchain technology is gaining traction as a platform for creating decentralized applications and facilitating cross-organizational processes~\cite{klinkmuller2019mining}.  However, extracting data to enable the analysis of process views from blockchains is notably challenging due to the decentralized nature of the database. The decentralized database applied in the blockchain stores transactional information triggered by smart contracts. These contracts contain the state and execution details of distinct cases~\cite{DBLP:journals/emisa/MendlingW18}. Retrieving data containing execution costs and the state set of activities enables process monitoring, compliance, and conformance checking through process mining. The workflow of process mining for the blockchain involves analyzing the smart contract and implementing the business process to identify transactions associated with task execution~\cite{DBLP:conf/bpm/MuhlbergerBCGL19}, as shown in Fig.~\ref{fig:blockchain}. In~\cite{DBLP:conf/bpm/MuhlbergerBCGL19}, the authors use the factory pattern to control the creation of process instances for blockchain.  In this pattern, a factory contract deploys a smart contract for each new process instance, thereby enacting the business process.
       
\begin{table}[!t]
\centering
\caption{Commonly Used Datasets in Process Mining.}
\label{tab:table1}
\renewcommand{\arraystretch}{1.0}
\scriptsize 
\begin{tabular*}{1\linewidth}{ccccccc}
\toprule
\textbf{Log Name / Author} & \textbf{Domain} &   \textbf{Year} & \textbf{Cases} &\textbf{Events}\\
\midrule
BPIC11&Healthcare&2011&1,143&150,291\\
BPIC12 & Finance & 2012 & 13,087&262,200\\
BPIC17 & Finance & 2017 & 21,861&714,198 \\
BPIC19 & Procurement & 2019 & 251,734 &1,595,923\\
BPIC20$_1$ & Reimbursement & 2020 & 6,886 &36,796 \\
BPIC20$_2$ & Reimbursement & 2020 & 10,500  &56,437 \\
BPIC20$_3$ & Reimbursement & 2020 & 2,099 &18,246 \\
BPIC20$_4$ & Reimbursement & 2020 & 6,449  &72151 \\
BPIC20$_5$ & Reimbursement & 2020 & 7,065 & 86,581 \\
Loan Application 1&Finance&2013&100&590\\
Loan Application 2&Finance&2013&70&420\\
Loan Application 3&Finance&2013&200&800\\
Loan Application 4&Finance&2013&105&630\\
RTFMP & Traffic & 2015 & 150,370 &561,470\\
SEPSIS & Healthcare & 2016 & 1,050 &15,214\\
Italian SW Co. Help Desk&Management&2017&4,580&21,348\\
Credit Requirement Logs&Finance&2017& 10,035&-\\
Doc Processing Logs&Management&2018& 18,352&-\\
Electronic Invoice Logs&Finance&2018&20135&-\\
CoSeLoG WABO&Public Services&2022&1434&8577\\
~\cite{8907303}&Logistics &2019&1450&47575\\
~\cite{9383956}&Logistics &2020&-&-\\
~\cite{10.1145/3462757.3466137}&Business Law&2021&4,795&266,834\\
\bottomrule
\end{tabular*}
\end{table}

\begin{table}[!t]
\centering
\caption{Fundamental Methods in Process Mining.}
\label{tab:table2}
\renewcommand{\arraystretch}{1.0}
\begin{tabular*}{0.9\linewidth}{ccc}
\toprule
\textbf{Process Mining Types} & \textbf{ Techniques / Algorithms}&\textbf{Papers} \\
\midrule
Discovery & $\alpha$-Algorithm &~\cite{1316839}\\
Discovery&Genetic Process Mining&~\cite{10.1007/11678564_18}\\
Discovery&Heuristic Mining&~\cite{f8cf09f3d2e243e8afdccaf7ea57ffd8}\\
Discovery&Region-Based Mining&~\cite{10.1007/978-3-540-85758-7_26}\\
Discovery&Inductive Mining&~\cite{10.1007/978-3-642-38697-8_17}\\
Conformance checking&Token Replay&~\cite{ROZINAT200864}\\
Conformance checking&Alignments&~\cite{97abafe6a895438ea808b3a6a7310d30}\\
\bottomrule
\end{tabular*}
\end{table}

\subsection{Datasets and Tools}
\subsubsection{Datasets}
Process mining assumes the existence of an event log in which each event refers to a case, an activity, and a point in time. An event log can be seen as a collection of cases and a case can be seen as a trace or sequence of events.
Event data may come from a wide variety of sources, including:
a database system, a comma-separated values (CSV) file or spreadsheet, a transaction log, a business suite/ERP system, a message log, or an open API providing data from websites or social media. Some papers have shared datasets used in experiments, which are also being widely utilized. The commonly used datasets from different application domains, such as healthcare, finance, procurement, reimbursement, traffic, management, public services, logistics, and business law in process mining are outlined in Tab. \ref{tab:table1}.

\subsubsection{Tools}
The fundamental methods in process mining are listed in Tab. \ref{tab:table2}. In detail, techniques used in process mining studies include different algorithms to discover process models ($\alpha$-algorithm, genetic process mining, heuristics mining, region-based mining, inductive mining, genetic mining, and inductive mining), and conformance checking (token replay, alignments). To analyze event logs with these process mining techniques and algorithms, a variety of tools are available.
ProM is an extensible framework that supports a wide variety of process mining techniques in the form of plug-ins. Disco consists of a licensed tool with a friendly visual interface for process models and easy functionality to apply multiple and variable filtering options in event logs. In addition, PM4py is a Python library that supports process mining algorithms in Python. It is open source and intended for use in both academic and industrial projects.

\subsection{Metrics and Criteria}
The four main quality criteria for evaluating the quality of process mining results are as follows: (1) Fitness: Fitness measures the ability of the model to reconstruct event logs. Metrics of fitness include error rate, accuracy, precision, recall, and other indicators. (2) Complexity: Complexity quantifies how difficult it is to understand a model. Complexity metrics encompass various factors, including size (measured by the number of nodes), control-flow complexity (CFC) (which accounts for the level of branching introduced by gateways in the model), and structuredness (the proportion of nodes situated within a block-structured single-entry single-exit fragment). (3) Precision: improving precision entails ensuring that the model does not suffer from underfitting and avoiding introducing events that are unrelated to the used event log. (4) Generalization: increasing generalization means avoiding model overfitting. An overfitting model can only explain individual cases in event logs. Therefore, an excellent process mining algorithm needs to strike a balance between overfitting and underfitting.

The evaluation of a process or organization's performance can be determined in various ways. Generally, three aspects of performance are recognized: time, cost, and quality. Each of these performance dimensions can have different Key Performance Indicators (KPIs) defined, such as: (1) Time dimension: Lead time or flow time is the complete duration from case creation to case completion. Service time refers to the duration spent on working on a case. Waiting time indicates the duration a case waits for an available resource. Synchronization time refers to the duration when an activity is not fully enabled and awaits an external trigger or another parallel branch. We need to control these indicators, which enables the process to be completed on time. (2) Cost dimension: Costs depend on the processing time and project requirements. The average resource utilization is often an essential performance indicator in most processes. (3) Quality dimension: In a process, quality refers to the indicators that measure the final service and results, and it needs to specify different indicators for different projects, such as customer satisfaction, product quality period, and repair rate.

In addition, the main criteria used for conformance checks are: (1) Adaptability, whether the model can extract events contained in the logs, (2) Generalizability, whether it can accept similar events related to previous log events and avoid overfitting, (3) Simplicity, the model should be as simple as possible for system evaluation, and (4) Precision, avoiding underfitting and the introduction of events irrelevant to the used logs~\cite{rosa2017business}. In addition to developing assessment criteria, the application of conformance checking, such as process monitoring, has also received extensive attention from researchers. The difference between them is that conformance checking involves analyzing complete logs after events have occurred, while process monitoring entails real-time monitoring using partial logs as events unfold.

\section{Intelligent Cross-Organizational Process Mining}
\begin{figure*}[!t]
    \centering
    \includegraphics[width=1\textwidth]{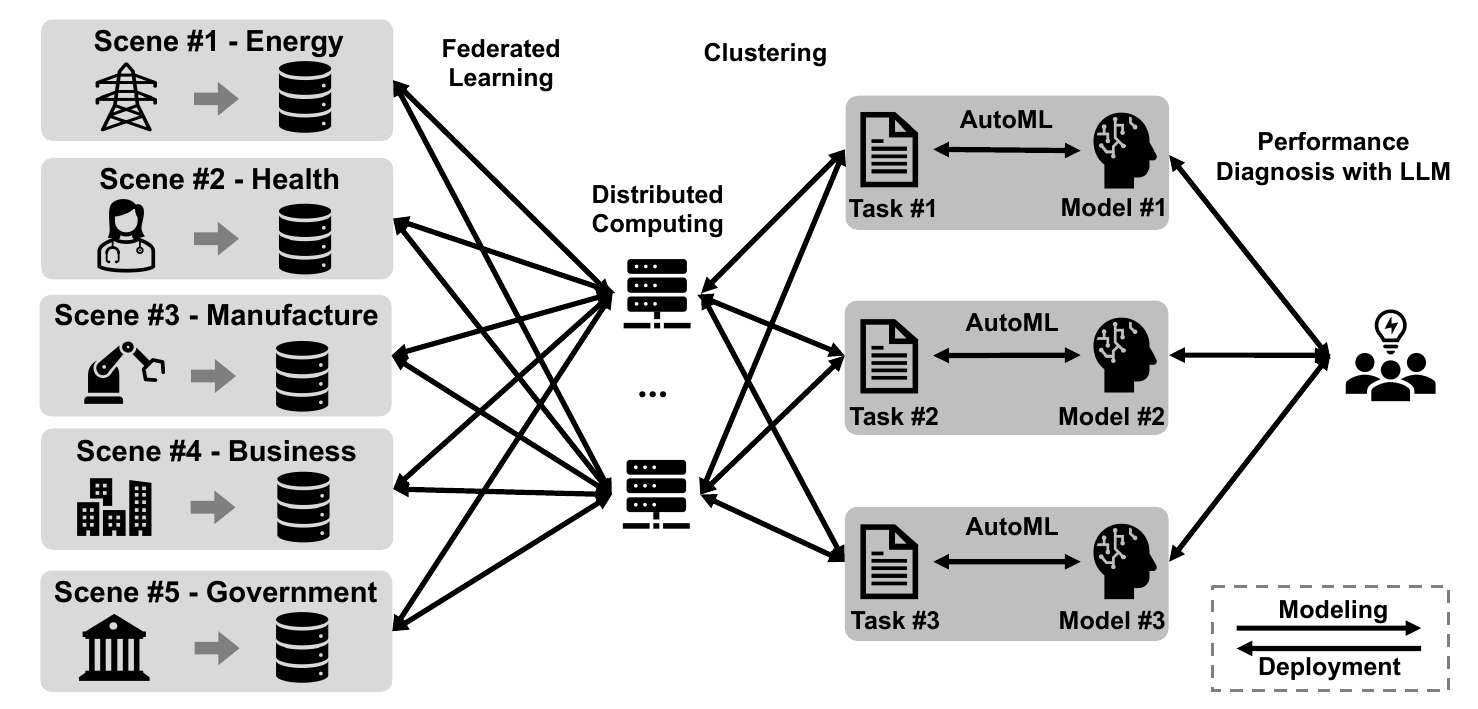}
    \caption{The workflow of the proposed \textbf{Intelligent Cross-organizational Process Mining}. Specifically, for different application scenarios, the system uses federated learning and distributed computing to improve data privacy, adaptability to cross-organizational event logs, and system efficiency. Next, clustering is applied to classify different sub-tasks by the event log's similarity and correlation. AutoML is then used to model each sub-task to further optimize the system's robustness to multi-task and adaptability to multi-model data. Finally, the system is combined with LLM for performance diagnosis and deployment.} 
    \label{fig:cross-ori}
\end{figure*}
In this section, we propose a new research direction, intelligent cross-organizational process mining, along with some initial solutions, as shown in Fig.~\ref{fig:cross-ori}. Currently, with the development of AI, an increasing number of process mining applications are incorporating intelligent technologies. As cross-organizational collaboration becomes more prevalent, AI-enabled cross-organizational process mining can be a hot topic. We will analyze it and provide initial solutions for cross-organizational process mining from three perspectives: data modality, model building, and downstream task.

\subsection{Data Modality}
In cross-organizational process mining, data modality refers to the types and sources of event logs from different organizations. These could include separate data from different organizations, structured and unstructured data, and real-time data streams~\cite{jabeen2023review}. The challenge lies in integrating and standardizing diverse data formats and modalities and ensuring privacy and security across organizational boundaries. Specifically, AI algorithms often differentiate modalities of log data based on their categories before building and constructing models accordingly. 

For instance, (1) Language-type logs, such as medical logs or work reports. In this case, natural language processing (NLP) methods are often considered a priority for model building~\cite{khurana2023natural}. Those methods are adept at parsing textual data and extracting contextual information from sentences. (2) Sequence-type logs, such as financial data or logistics sequences. Time series analysis is effective, as it can capture trends, cyclic patterns, and seasonal features~\cite{wen2022transformers}. More deeply, for spatio-temporal data, as with general time series data, we will continue to consider features and patterns in both the temporal and spatial dimensions on a temporal basis~\cite{wang2020deep, yang2024survey}. (3) Tabular-type logs, such as bank statements, factory processes, and hospital procedure forms. We usually apply some machine learning-based methods to extract the features between records (rows) and attributes (columns)~\cite{borisov2022deep}. In the above three cases, the model building is driven by an independent data modality. This tailored approach ensures that the models are suitable for the specific application scenarios and nuances of the data modalities they are dealing with. Therefore, enhance the accuracy and effectiveness of the outcomes.

Furthermore, different organizations may have diverse data modalities, and where data privacy between organizations is a paramount concern, the adoption of federated learning is indeed a strategic approach~\cite{khan2021cross}. Federated learning is a machine learning technique that allows for the construction of a common, collaborative model while keeping the training data localized~\cite{zhang2021survey}. In essence, the model is trained across multiple decentralized devices or organizations without exchanging the data itself. This approach not only preserves data privacy but also leverages diverse data sources to create a more robust and comprehensive model~\cite{van2021federated}. Specifically, this method offers several key advantages in the context of cross-organizational process mining: (1) Data privacy and security: By keeping data localized and not sharing it with other organizations or a central server, federated learning ensures the privacy and confidentiality of organizational data. (2) Handling of diverse data modalities: Federated learning can accommodate different data types and structures from various organizations. Each organization can train a local model on its own data modality and realize efficient distributed computing, and these local models can then contribute to a global model. (3) Collaborative learning: Federated learning enables collaborative model building, where all participants can benefit from shared insights without compromising their proprietary data.

\subsection{Model Building}
The intelligence in process mining indeed hinges on the model building. In specialized AI domains, the selection of suitable models often varies according to the concrete sub-tasks and data modalities involved. For instance, decision trees are effective for handling discrete or categorical data, often found in tabular formats such as spreadsheets and databases~\cite{maimon2014data}. Markov models are adept at handling sequential data~\cite{rabiner1986introduction}. RNNs, and their variants, like LSTMs and transformers, are designed for sequential data with dependencies over time, making them ideal for language data in process mining~\cite{lipton2015critical}.

Nowadays, the increasing demand for process mining across various application scenarios, each with its own distinct categories and tasks, indeed makes the traditional approach of individually modeling for each task inefficient and less robust. This is why AutoML has become extremely valuable. AutoML refers to the process of automating the end-to-end process of applying machine learning to real-world problems~\cite{he2021automl}. In the context of process mining, AutoML can significantly streamline and enhance the efficiency and effectiveness of model development. There are some benefits in applying AutoML to process mining. (1) Efficiency in model development: AutoML automates the selection, composition, and parameterization of AI models. This reduces the time and resources required to develop models for each specific sub-task in process mining. (2) Increased robustness: By automatically selecting the best models and tuning parameters, AutoML can achieve more robust performance than manually developed models. (3) Accessibility: AutoML can make advanced process mining techniques more accessible to organizations that may not have extensive expertise in AI. (4) Scalability: AutoML can quickly adapt, providing solutions for new types of event logs, tasks, or changing business environments without the need for extensive reprogramming or manual model adjustment. (5) Continuous improvement: AutoML systems can continuously learn and improve from new data, ensuring that the models remain up-to-date and effective over time. 

Overall, implementing AutoML in process mining involves using automated systems to handle model building and training processes, such as data preprocessing, feature engineering, model selection, hyperparameter tuning, and model evaluation. It will save time and resources and also potentially uncover novel insights and patterns that manual modeling processes might overlook.

\subsection{Downstream Task}
For process mining, especially for the deployment process, different sub-tasks are handled based on various application scenarios, and AI models are constructed to produce different outputs according to these sub-tasks. Common tasks in process mining include: (1) Prediction: It uses historical data to estimate future events, such as the duration until completion of a process or predicting the likelihood of various possible outcomes~\cite{zhao2021event}. (2) Anomaly detection: AI models for anomaly detection identify patterns that deviate from the norm, which can indicate potential issues or areas of concern in a process~\cite{pang2021deep,yang2023dcdetector}. (3) Classification: Classification models process instances based on defined criteria, which can be crucial to ensuring that processes adhere to standards or meet performance benchmarks~\cite{minaee2021deep}. Compliance checks and performance analysis fall under this category. (4) Attribution: This involves identifying the causes of issues or inefficiencies in a process and finding the root causes of deviations. Attribution models in process mining analyze data to trace back and understand the contributing factors to a particular outcome or anomaly~\cite{sole2017survey}. Overall, each of these tasks requires a specialized model tailored to the specific event data and the objective of the analysis. The effectiveness of AI in process mining depends largely on the correct application of these models to their respective tasks.

When dealing with numerous repetitive or similar task scenarios, adopting a platform-centric approach is indeed crucial to intelligent cross-organizational process mining. The core of it is clustering, which involves grouping similar tasks together in application scenarios. This strategy offers several significant benefits. (1) Efficiency optimization: Clustering similar tasks enables the development of generalized solutions that can be efficiently applied across multiple scenarios. It reduces redundancy in developing individual solutions for each sub-task, thereby improving efficiency. (2) Enhanced performance: By aggregating similar tasks, the log data becomes more substantial and diverse, leading to more robust models. It can improve the accuracy and reliability of process mining outputs. (3) Scalability: A platform-centric approach with clustering at its core allows for easier scalability. As new tasks or scenarios emerge, they can be quickly integrated into existing clusters, making the system adaptable and responsive to changing needs. (4) Knowledge transfer: Clustering enables the transfer of insights and learning from one task to another within the same cluster. It can lead to continuous improvement and innovation within grouped task scenarios. (5) Customization and standardization: While clustering allows for standardizing approaches and solutions, it still accommodates customization within each cluster, ensuring that specific needs and nuances of individual tasks are addressed. (6) Resource allocation: By identifying and clustering similar tasks, resources can be allocated more effectively, focusing efforts where they are most needed and avoiding duplication.

\section{Discussion}
\subsection{Current Challenges}
Enhancing process intelligence is achievable through the strategic utilization of data collected from various enterprises. Investigating strategies for utilizing data from a collective of organizations to address a diverse set of problems, even those that may not be commonly shared, is of significant value for extensive research and exploration. The industry prioritizes platform capabilities over the needs of individual organizations. By focusing on solutions that address challenges across all organizations, not only can problems be solved more comprehensively, but the solutions are also likely to be of higher quality than those tailored for a single entity. Therefore, process mining should adopt a cross-organizational approach. Cross-organization process intelligence holds significant intrinsic value, yet it is not without substantial challenges. In this section, we will delve into key challenges observed from an industrial perspective.

\subsubsection{Data Privacy} 
Organizations, such as enterprises or governments, typically place a high premium on securing their data within their domains to prevent external dissemination. 
The efficacy of process intelligence algorithms that rely on leveraging data can be compromised by concerns related to data privacy. However, it is essential to recognize that leveraging data across enterprises has the potential to significantly enhance techniques, such as the development of robust modeling. This balance between data confinement and strategic utilization across organizational boundaries can unlock new opportunities for advancements in modeling and other data-driven methodologies.

\subsubsection{Data Inconsistency} 
Various organizations exhibit disparities in data formats, data volumes, data distributions, and even the interpretation of data labels, posing additional challenges for process intelligence algorithms. Take data distribution as an example. Given that clients commonly collect data from distinct environments, heterogeneous data distribution becomes a serious challenge in federated learning, which has a negative impact on the training of federated learning~\cite{DBLP:journals/corr/abs-2310-07171,li2023federated}.

\subsubsection{Customizable Computation}
Recognizing diverse organizations' requirements, particularly in areas such as different precision-recall tradeoffs, is important. Given that enterprises may have distinct demands, providing customizable computation capabilities becomes crucial to ensuring flexibility and adaptability to a wide array of organizational needs. This approach allows for a tailored and responsive solution that aligns with the unique preferences and specifications of each enterprise.

\subsubsection{Long Tail Distribution of Applications} 
Diverse organizations operate with distinct processes, underscoring the intricacies of tailoring solutions to specific operational contexts. For instance, the challenges faced by the food production industry vastly differ from those in the automotive sector. Given the wide array of unique domains, numerous and varied problems arise. Many of these problems may not be shared across multiple organizations, resulting in sparse data for each specific issue. In addressing this diversity, it is essential to pinpoint commonalities and insights across these applications. Utilizing these insights as guidance, the algorithm and system design can be tailored to better accommodate the distinct nuances of each operational context while covering a wide range of applications.

\subsubsection{Explainability}
Industry process intelligence aims to provide valuable insights for organizations. It is crucial for an organization to comprehend how this intelligence is computed. Besides, the explainability of intelligent models can help non-experts understand and apply AI-based systems. Models with good explainability also allow users to better perform problem attribution and process reviews.

\subsubsection{Other Potential Challenges}
(1) Large-scale data: The imperative for processing extensive datasets becomes prominent due to the proliferation of organizations, coupled with the escalating volume of data within each organization. Furthermore, with the passage of time, there is a continual increase in the generation of data. (2) No/Missing labels: Several reasons present challenges in collecting data, particularly when it comes to obtaining labeled datasets. Organizations may not always be cooperative, and for certain tasks, it may be impossible to mandate organizations to provide labels. Organizations that have recently initiated specific applications may also encounter challenges due to the limited availability of data. (3) Rising demand for advanced AI techniques: With the increasing popularity of AI techniques, organizations are seeking a more intelligent understanding of their data. This necessitates support for advanced methods such as machine learning, including the integration of large language models (LLMs). In the context of data privacy concerns, the customization of LLMs for individual organizations becomes exceptionally important.

\subsection{Outlook and Future Opportunities}
In this section, we explore opportunities within the research field of cross-organization process intelligence, focusing on avenues that have the potential to address the challenges discussed above.

\subsubsection{Federated Learning for Cross-organizational Process Mining} As we discussed in the previous section, the adoption of federated learning allows the creation of global models without exposing organizations' data. This approach prevents data leakage issues, enhances model quality for individual organizations, and benefits organizations with limited or no data through a shared global model.

\subsubsection{AutoML for Cross-organizational Process Mining} AutoML emerges as a comprehensive solution for streamlining tasks ranging from feature processing to model selection, training, and deployment. In the context of diverse organizational needs, applications may vary, necessitating the construction of organization-specific models. Managing a multitude of models, each corresponding to a different organization, can prove costly and challenging. AutoML addresses this by adeptly adapting to organization-specific configurations, offering a scalable solution that accommodates diverse requirements. AutoML allows us to concentrate on developing a universal AutoML backbone system. This framework can then be tailored by different organizations to create customized AutoML pipelines suited to their specific tasks. Moreover, AutoML contributes to privacy preservation. For organizations keen on safeguarding their data within their own confines, AutoML offers a means to achieve machine learning-based intelligence without exporting any sensitive or aggregated data externally. Through the training, validation, and serving of models within their realms, organizations can confidently maintain data privacy while still benefiting from the insights provided by AutoML.

\subsubsection{Explainable Model for Cross-organizational Process Mining}
Prior to organizations entrusting and adopting process intelligence techniques, a fundamental requirement is the ability to comprehend and validate the information generated. Consequently, the provision of explainable results becomes a critical factor in establishing trust and facilitating widespread adoption, particularly in cross-organizational scenarios. (1) Explainable results for trust: Implementing process intelligence should include methodologies that yield results in an explainable manner. By making the results interpretable and accessible, stakeholders can gain confidence in the reliability and accuracy of the process intelligence system. (2) Facilitating cross-organizational support: When process intelligence results are easily understandable, it simplifies communication and collaboration between different entities. Organizations can grasp the insights without requiring extensive interaction with the process intelligence provider, streamlining the adoption process and enabling efficient utilization of shared information. In essence, ensuring the explainability of process intelligence not only aids comprehension but also enhances the overall accessibility and utility of process intelligence, paving the way for broader acceptance and integration into various organizational contexts.

\subsubsection{Foundation Models for Cross-organizational Process Mining}
LLMs are becoming increasingly vital, particularly when trained to comprehend domain-specific knowledge tailored to an organization's needs. Research in the field of LLMs is geared towards addressing two significant aspects: (1) Local serving of LLMs: In the quest for privacy and data security, there is a growing emphasis on serving LLMs locally within an organization. This approach mitigates the risk of sensitive data leakage to third-party providers. Localized serving of LLMs ensures that the valuable information processed by these models remains within organizational boundaries, adhering to stringent privacy and security standards. (2) Domain-specific LLM training: Recognizing the unique requirements of different organizations, research is directed towards training LLMs to possess a deep understanding of specific domain knowledge relevant to each entity. By tailoring LLMs to comprehend the intricacies of an organization's domain, these models can provide more accurate, context-aware, and valuable insights. This customization ensures that the LLM is not only linguistically proficient but also contextually aligned with the specific needs and nuances of the organization. In essence, the dual focus on local serving for enhanced data security and domain-specific training for contextual understanding reflects the evolving landscape of LLM research. These advancements contribute to the development of more powerful, secure, and organizationally aligned language models, offering a sophisticated tool for leveraging language processing capabilities while safeguarding sensitive data.

\subsubsection{Scalable Computation for Cross-organizational Process Mining}
The scale of data today presents opportunities for leveraging distributed computing methodologies to enhance computational capabilities. Offering distributed computation as a service introduces a transformative approach, eliminating the need for individual organizations to invest in expensive resources for distributed computing. By providing distributed computation capabilities as a service, organizations can harness the power of large-scale data processing without shouldering the financial burden of setting up and maintaining dedicated infrastructure. This shift allows them to focus resources on their core competencies, while still benefiting from the scalability and efficiency offered by distributed computing.

\section{Conclusion}
In this survey, we explored the landscape of modern intelligent processing within the realm of cross-organizational process mining, emphasizing the integration of AI technologies to bolster efficiency and decision-making in complex, multi-organizational environments. Our examination brought to light the critical challenges of data privacy, computational complexity, and data inconsistency, which are pivotal to advancing this field. We also proposed a comprehensive workflow tailored for intelligent cross-organizational process mining, underscoring the importance of scalable computation and the development of explainable models. These elements are essential for fostering trust and facilitating the adoption of AI-driven solutions in multi-organizational settings. By synthesizing existing research and identifying gaps, this survey aims to guide both practitioners and researchers towards new directions in the intelligent multi-organizational process mining domain. The insights and frameworks discussed herein lay the groundwork for future innovations and exploration, potentially setting new standards and practices in the field. Our findings suggest a promising horizon where enhanced AI integration can address current limitations and unlock new opportunities for process mining across organizations.

\bibliographystyle{IEEEtran}
\bibliography{7_ref}

\end{document}